%% file: main.tex
\pdfoutput=1

\documentclass[11pt]{article}

\usepackage[]{acl}

\usepackage{fnpos}
\usepackage{times}
\usepackage{latexsym}
\usepackage{multirow}
\usepackage{graphicx}
\usepackage{fourier} 
\usepackage{array}
\usepackage{amssymb}
\usepackage{makecell}
\usepackage[T1]{fontenc}

\usepackage[utf8]{inputenc}

\usepackage{microtype}

\usepackage{inconsolata}
\usepackage{listings}

\usepackage{multirow}
\usepackage{graphicx}
\usepackage{booktabs}
\usepackage{amsmath}
\usepackage{amssymb}
\usepackage[normalem]{ulem}
\useunder{\uline}{\ul}{}
\usepackage{arydshln}
\usepackage[table]{xcolor}    
\definecolor{warningcolor}{RGB}{250,35,64}
\definecolor{myLightBlue}{RGB}{200,230,255}
\definecolor{myBlue}{RGB}{0, 102, 202}

\definecolor{myGreen}{RGB}{0, 153, 102}   
\definecolor{myRed}{RGB}{204, 0, 0}   
\definecolor{aqua}{rgb}{0.0, 1.0, 1.0}
\definecolor{orange}{RGB}{244,164,71}
\definecolor{myOrange}{RGB}{244, 130, 50}
\definecolor{yellow(ryb)}{RGB}{249,195,0}

\usepackage{colortbl}
\usepackage{amssymb}
\usepackage{pifont}
\usepackage{float}
\usepackage[most]{tcolorbox}    
\tcbuselibrary{breakable}       

\newtcolorbox{prompt}[2][]{
    colback=gray!20,
    colframe=white,
    fonttitle=\bfseries\small,
    boxrule=0.4mm,
    fontupper=\small, 
    fontlower=\small,
    coltitle=white,
    title=#2,
    #1,breakable
}

\definecolor{codebg}{RGB}{248,248,248}
\definecolor{commentgray}{gray}{0.4}
\definecolor{keywordblue}{rgb}{0.26, 0.26, 0.9}

\lstdefinestyle{pythonstyle}{
    backgroundcolor=\color{codebg},
    language=Python,
    basicstyle=\ttfamily\scriptsize,
    keywordstyle=\color{keywordblue}\bfseries,
    commentstyle=\color{commentgray}\itshape,
    stringstyle=\color{orange},
    showstringspaces=false,
    breaklines=true,
    frame=single,
    framerule=0.5pt,
    xleftmargin=1em,
    xrightmargin=1em,
    tabsize=4
}

\title{Speak \& Spell: LLM-Driven Controllable Phonetic Error Augmentation for
Robust Dialogue State Tracking
}

\author{Jihyun Lee$^1$, Solee Im$^3$\thanks{This work was conducted at POSTECH.}, Wonjun Lee$^2$, Gary Geunbae Lee$^{1,2}$ \\
  $^1$Graduate School of Artificial Intelligence, POSTECH, Republic of Korea\\
  $^2$Department of Computer Science and Engineering, POSTECH, Republic of Korea\\
  $^3$KT\\
  \texttt{\{jihyunlee, lee1jun, gblee\}@postech.ac.kr} \\
  \texttt{\{solee.im\}@kt.com}
}

\begin{document}
\maketitle
\begin{abstract}
Dialogue State Tracking (DST) is a key part of task-oriented dialogue systems, identifying important information in conversations. However, its accuracy drops significantly in spoken dialogue environments due to named entity errors from Automatic Speech Recognition (ASR) systems. We introduce a simple yet effective data augmentation method that targets those entities to improve the robustness of DST model. Our novel method can control the placement of errors using keyword-highlighted prompts while introducing phonetically similar errors. As a result, our method generated sufficient error patterns on keywords, leading to improved accuracy in noised and low-accuracy ASR environments.
\end{abstract}
\input{introduction}

\input{method}

\input{experiment}
\input{analysis}

\vspace{-4pt}
\section{Conclusion}
\vspace{-4pt}

We propose a novel data augmentation method tailored for DST tasks that ensures sufficient error patterns in both key phrases and overall text. By leveraging LLMs for their controlled text generation capabilities, we strategically place errors within key phrases. Our method demonstrates substantially improved robustness in DST by generating diverse, plausible keyword errors. Error case analysis reveals that keyword augmentation significantly enhances robustness against ASR errors.  As the pioneering research in leveraging LLMs for generating ASR errors, we hope this work lays a strong foundation for future phonetic-based augmentation research.

\section*{Limitations}

Through detailed error analysis, we identified a trade-off introduced by our keyword-focused phonetic augmentation strategy. While the augmentation helps the model become more robust to noisy slot expressions—leading to substantial reductions in "Wrong" errors—it also increases the model’s sensitivity to phonetic variations. As a result, we observed cases where the model hallucinates slot values that were not actually mentioned, thereby increasing the number of "Spurious" errors. This hallucination effect represents a key limitation of our method. We attribute it to the model's repeated exposure to noisy keywords, which may cause it to overgeneralize phonetic cues as valid slot mentions. As a direction for future work, we plan to incorporate an auxiliary loss term for slot presence prediction~\cite{trippy,kim2019efficient} to help the model better distinguish between mentioned and unmentioned slots and mitigate this side effect.

\section*{Ethical Considerations}
Our phonetic augmentation method, while effective for simulating ASR-style errors, may raise several ethical concerns. One such concern is the potential for accent bias, wherein phonetic transformations may disproportionately reflect majority or standard pronunciations, thereby marginalizing regional or minority accents. Another concern is the inadvertent corruption of proper names, particularly those that are less common or culturally specific, which could lead to misrepresentation or reduced inclusivity. We acknowledge these risks and emphasize that our method relies on LLMs trained on diverse and large-scale corpora. As such, the phonetic errors generated are likely to reflect dominant patterns present in mainstream ASR systems, rather than rare or region-specific variations. Nonetheless, we recognize the importance of fairness and inclusivity in language technologies and believe that future work should explore augmentation strategies that are more sensitive to accent and cultural variability.

\section*{Acknowledgements}
This research was supported by the Culture, Sports and Tourism R\&D Program through the Korea Creative Content Agency funded by the Ministry of Culture, Sports and Tourism in 2025 (Project No. RS-2025-02413038, 45\%),
the IITP-ITRC (Information Technology Research Center) Program funded by the Ministry of Science and ICT (Project No. IITP-2025-RS-2024-00437866, 45\%),
and the IITP Artificial Intelligence Graduate School Program (POSTECH) funded by the Ministry of Science and ICT (Project No. RS-2019-II191906, 10\%).

\bibliography{anthology,custom, reference}

\clearpage
\appendix
\input{appendix}

\end{document}

%% file: introduction.tex
\section{Introduction}

\begin{figure*}[t]
  \centering
  \includegraphics[width=\textwidth]{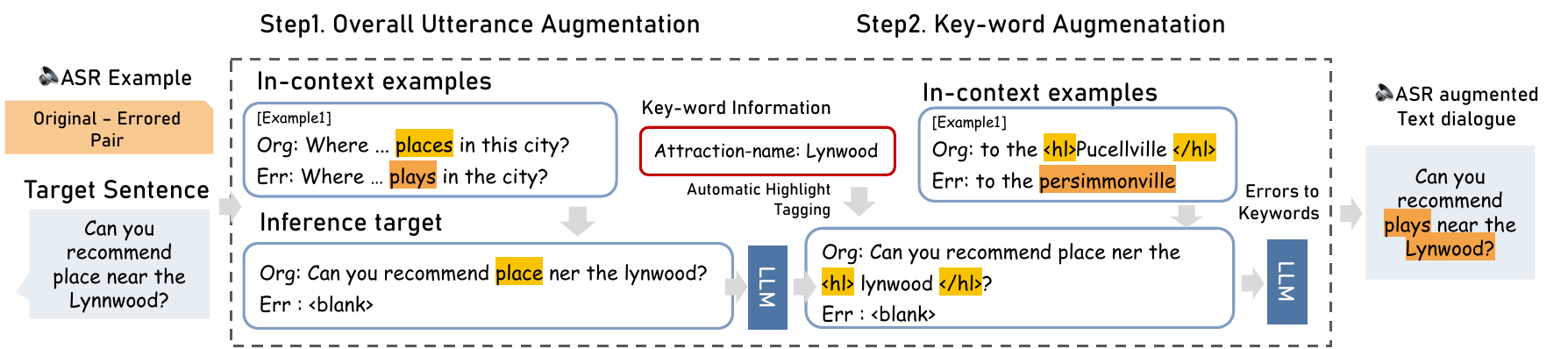}
\caption{
Illustration of the EPA process. 
Step~1 introduces overall ASR-like distortions using in-context examples of original–errored pairs, while Step~2 applies keyword-specific phonetic corruptions guided by highlighted slot spans. 
Combining both steps generates realistic and diverse ASR-style errors for robust DST training.
}
   \label{fig:method}
   \vspace{-13pt}
\end{figure*}

Task-oriented dialogue systems (TODs) assist users in achieving specific objectives through conversations and are used in various sectors, including customer service and hotel reservations. A crucial component of these systems is Dialogue State Tracking (DST), which extracts vital information from conversations in a slot-value format (e.g., hotel-name: Claire Hotel). This information is essential for querying databases and generating responses \cite{young2013pomdp}.

However, DST models face significant challenges in spoken dialogue environments, where user utterances are converted into text by automatic speech recognition (ASR) \cite{pal2020modeling, kim2021robust, yoon2023adapting}. Notably, \citet{dstc11} observed a drastic reduction in model accuracy from 41.6\% to 23.6\% in such environments. This decline is primarily due to ASR errors, which frequently misrecognize named entities—a key target in DST \cite{Nechaev2021}.

To address ASR inaccuracies, data augmentation has emerged as a viable, cost-efficient strategy. Existing text augmentation methods, such as word swapping \cite{eda} and back translation \cite{BT}, do not maintain audio similarity with the original text, leading to discrepancies with ASR error patterns. To bridge this gap, \citet{sharma2020improving} and \citet{jacqmin2023olisia} synthesized audio from text with text-to-speech (TTS) model \cite{taco} and processed it through ASR, while \citet{sc-BERT} and \citet{sc-kfold} employed translation model structure to introduce ASR-like errors directly into texts. \citet{huang2020learning} further leveraged word confusion network (WCN) representations for data augmentation to generate acoustically consistent errors. 

Despite these advancements, prior methods often fail to provide sufficient error for DST model training. Accurately identifying key terms is vital for DST performance; thus, models need to be trained on a broad spectrum of ASR-errored keywords. Unfortunately, many current strategies do not ensure that errors are positioned within critical keywords, often generating trivial examples by altering non-essential words such as random words \cite{eda} or sentence structure \cite{BT}. This oversight results in sub-optimal performance against ASR errors.

To address these limitations, we introduce Error Positioning Augmentation (EPA), a straightforward yet effective method that ensures sufficient errors in keywords. Our method leverages large language models (LLMs) \cite{openai, touvron2023llama, zhang2022opt}, which have demonstrated impressive capabilities in semantic augmentation \cite{semaug1, semaug2} and precise text generation control \cite{control1, control2}. Despite their strengths, LLMs' potential for phonetic augmentation remains largely unexplored.

In our method, we utilize in-context learning~\cite{incontext} with phonetically similar examples to introduce general ASR errors and devise a highlighting method to explicitly localize the error to a target span. Surprisingly, without requiring extensive domain-specific user speech data, a publicly available audio dataset and a small set of in-context examples (fewer than 10 samples) are sufficient to generate a wide variety of ASR-errored keywords for DST. This significantly simplifies the error generation process.

In the experiment, to reflect diverse real-world conditions, we evaluated EPA under four ASR environments: a low-accuracy ASR system, noisy audio with café and traffic background, a paraphrased input setting where users naturally rephrased transcriptions, and a high-accuracy ASR system. In these experiments,  EPA significantly improved model robustness, increasing accuracy from 45.76\% to 51.12\% with high keyword diversity (95.4\%), surpassing the previous best-performing model. Our analysis suggests that this improvement is primarily driven by keyword-level augmentation, which effectively mitigates errors in ASR-affected values.

%% file: method.tex
\vspace{-2pt}
\section{Method}
\vspace{-3pt}

\subsection{Notation}
Before detailing each step, we first clarify the notation. Dialogue context from turn 1 to $t$ is denoted as $D_t$=$\{(s_1, u_1 ),..., ( s_t, u_t )\}$ where $s$ denotes for system and $u$ for user utterance. DST model predicts the dialogue state (also called belief state) $B_t$ given $D_t$. $B_t$ is composed with slot $sl$ and value $v$  pairs, denoted as $B_t$ = $\{( {sl}_1, v_1 ),..., ( {sl}_J, v_J )\}$ , where ${sl}_j$ and $v_j$ is $j$-th slot name and value. $J$ is the total number of slots.

\subsection{Step 1: ASR Error for Overall Utterance}
\label{sec:method_step1}
In this step, we augmented the overall utterance by introducing general ASR errors. We began by constructing example sets for in-context learning, utilizing an open-source audio dataset \cite{ardila-etal-2020-common}. From this dataset, we randomly selected 300 hours of audio along with their corresponding gold transcripts ($g$) and transcribed the audio using an off-the-shelf ASR model (e.g., Whisper-base \cite{whisper}) to obtain the erroneous transcriptions ($e$). We denote this example dataset as $DB = \{(g_1, e_1), \ldots, (g_I, e_I)\}$.

Next, we inject errors into $u$ by prompting the LLM with in-context examples. We retrieved ($g$, $e$) pairs from the database ($DB$) based on phonetic similarity between $u$ and $g$ (Figure~\ref{fig:method}, Step 1). To compute phonetic similarity, we converted the characters of both $u$ and $g$ into phonemes using the International Phonetic Alphabet (IPA), and calculated similarity using a frequency-based retrieval \footnote{We used BM25\cite{bm25}, a retrieval model based on term frequency. While neural retrievers (e.g., DPR\cite{DPR}) could be applied, we opted for a frequency-based method, as neural models tend to capture semantic similarity.}. After selecting the top-$k$ ($g$, $e$) pairs, we concatenated the instruction, in-context examples, and $u$ into a single prompt and provided it to the LLM. This process results in the overall ASR-errored user utterance, denoted as $\dot{u}$. Concretely, $\dot{u}$ can be obtained by 
 \begin{equation}
    \dot{u_t} = LLM(Inst_1 \oplus (g_1, e_1)\cdots(g_k, e_k)\oplus u_t)
\end{equation}
where $\oplus$ denotes concatenation, and we set $k=3$ throughout our experiments.  Retrieved examples are provided in Appendix~\ref{app:EPA_retreive}.

\subsection{Step 2: ASR Error for Keywords}
\label{sec:method_step2}

While Step 1 introduces general ASR-style errors into $u$, it does not ensure sufficient error diversity in keyword tokens. To construct a more effective training dataset, we explicitly generate keyword-focused ASR errors in Step 2 (Figure~\ref{fig:method}). In this step, we highlight the keywords in $\dot{u}$ using the \texttt{<hl>} tag and instruct the LLM to inject errors specifically within the highlighted spans. For the DST task, we treat dialogue state values ($v$) as keywords, although the definition of a keyword may vary depending on the task. To facilitate this process, we provide a few examples that illustrate how values within \texttt{<hl>} tags are intended to be modified during augmentation. Given these instructions and examples, the LLM generates an augmented utterance $\ddot{u}$ that includes both general and keyword-specific ASR errors. Formally, we obtain $\ddot{u}_t$ as follows: 
\begin{equation}
    \ddot{u_t} = LLM(Inst_2 \oplus ({g}_0, {e}_0)\cdots({g}_k, {e}_k)\oplus \dot{u}_t).
\end{equation}
The used prompts are provided in Appendix~\ref{app:prompt}.

\subsection{Examples of EPA}
\input{table/table_examples}
Table~\ref{tab:EPA_examples} shows examples of ASR errors generated by EPA. We have highlighted utterance level overall errors in \colorbox{yellow(ryb)}{yellow} and keyword-specific errors in \colorbox{orange}{orange}. For instance, in Row 1, the model introduces a keyword-level error (bailey’s → baley’s) as well as an additional phonetically plausible insertion (peas), simulating realistic ASR noise. Further examples can be found in Appendix~\ref{app:asr-examples}.

%% file: table/table_examples.tex
\begin{table}[t]
\resizebox{\columnwidth}{!}{%
\begin{tabular}{c|l|l}
\hline
            Idx &    Method & \multicolumn{1}{c}{Examples}                           \\ \hline
\multirow{2}{*}{1} & Original & Tuesday, going to bailey's crossroads please.           \\
                   & +EPA & Tuesday, going to \colorbox{orange}{baley's} crossroads, \colorbox{yellow(ryb)}{peas}.             \\ \hline            
\multirow{2}{*}{1} & Original & I'd like to find a vegetarian restaurant, if possible.  \\
                   & +EPA & I'd \colorbox{yellow(ryb)}{hike} to find a \colorbox{orange}{veggie tarian} \colorbox{yellow(ryb)}{restroom}, if possible. \\ \hline
\multirow{2}{*}{3} & Original & I am going to auburn.                                   \\
                   & +EPA & I am \colorbox{yellow(ryb)}{flowing} to \colorbox{orange}{auburng}.                                \\ \hline

\multirow{2}{*}{4} & Original & Hi! Could you find me a train to loris on thursday?     \\
                   & +EPA & \colorbox{yellow(ryb)}{Oh}! Could you find me a \colorbox{yellow(ryb)}{trai} to \colorbox{orange}{lorri} on \colorbox{orange}{thursdae}?      \\ \hline
\multirow{2}{*}{5} & Original & Ashby is my destination.                                \\
                   & +EPA & \colorbox{orange}{Ashy's} my \colorbox{yellow(ryb)}{desity}.                                       \\ \hline
\end{tabular}%
}
\caption{Examples of ASR errors from EPA. }
\label{tab:EPA_examples}
\vspace{-15pt}
\end{table}

%% file: experiment.tex
\vspace{-2pt}
\section{Experiments}
\vspace{-3pt}
\input{table/table_main}
\subsection{Experimental Setup}
\noindent
\textbf{Dataset.}  The DSTC11 dataset \cite{dstc11}, an audio version of MultiWOZ 2.1 \cite{woz2.1}, comprises 8,000 dialogues for training, 1,000 for validation, and 1,000 for testing. To enhance generalization, we conducted experiments across four distinct ASR environments, characterized by Word Error Rate (WER) and noise levels: (1) a low accuracy ASR model (WER > 0.03), (2) a café and traffic noised audio, (3) a paraphrased setting where users naturally paraphrased the transcriptions, and (4) a high accuracy ASR model.

\noindent
\textbf{Metrics.} For overall performance evaluation, we used joint goal accuracy (JGA), which requires all slot-value pairs to match the gold label. We also reported named entity accuracy (N-acc), the average accuracy across named entity slots.

\noindent
\textbf{Compared methods.}
We compared our method with two established approaches: text-based augmentations, \textbf{AEDA} \cite{aeda}, \textbf{EDA} \cite{eda}, and Back Translation (\textbf{BT}) \cite{BT}, and audio-aware augmentation methods, using synthesized audio (\textbf{TTS-ASR}) and translation model structure (\textbf{ASR-translation}). Lastly, we included \textbf{Olisia} \cite{jacqmin2023olisia}, the top-ranked method in the DSTC11 competition.

\noindent
\textbf{Models.} For performing EPA, we used diverse types of LLMs, including GPT-3.5\cite{openai}, \textsc{LLaMA2-7B}\cite{touvron2023llama} and OPT-6.7B\cite{zhang2022opt}.
For the DST task, we fine-tuned a T5-base\cite{t5} model. Further details about the experimental settings are provided in Appendix~\ref{app:exp_setting}.



%% file: table/table_main.tex
\begin{table*}[t!]
\centering

\resizebox{\textwidth}{!}{%
\begin{tabular}{lcccc|rr|rr|rr|rr}
\toprule
\multicolumn{1}{c}{\multirow{2}{*}{Method}} &
  \multicolumn{4}{c|}{Features} &
  \multicolumn{2}{l|}{Low-acc ASR} &
  \multicolumn{2}{l|}{Noised Aud.} &
  \multicolumn{2}{l|}{Paraphrased} &
  \multicolumn{2}{l}{High-acc ASR} \\ \cline{2-13} 
\multicolumn{1}{c}{} & 
  Aud. &
  Utt-aug &
  Key-aug &
  LLM &
  JGA &
  N-Acc &
  JGA &
  N-Acc &
  JGA &
  N-Acc &
  JGA &
  N-Acc \\\hline
Baseline       & - & - & - & - & 29.88 & 45.76 & 29.70  & 46.77 & 28.92 & 48.79 & 34.87 & 52.07 \\\hline
AEDA \cite{aeda}        & - & \checkmark & - & - & 29.90  & 46.46 & 29.74 & 47.48 & 29.12 & 48.86 & 34.94 & 52.32 \\
EDA \cite{eda}           & - & \checkmark & - & - & 29.22 & 47.65 & 28.70  & 49.51 & 28.08 & 49.99 & 33.68 & 53.78 \\
BT \cite{BT}             & - & \checkmark & - & - & 31.69 & 49.17 & 31.26 & 50.98 & 29.90  & 51.73 & 36.27 & 54.81 \\
TTS-ASR   & \checkmark & \checkmark & - & - & 29.94 & 46.34 & 29.99 & 47.37 & 29.08 & 48.88 & 35.07 & 52.03 \\
ASR-translation         & \checkmark & \checkmark & - & - & 30.40  & 47.65 & 30.14 & 48.45 & 29.54 & 50.38 & 35.25 & 53.66 \\ \hdashline

EPA (Opt 6.7B) & \checkmark & \checkmark & \checkmark & \checkmark & 31.82 & 50.73 & 32.03 & 51.92 & 29.57 & 52.49 & \textbf{37.05} & 55.78 \\
  
\quad \small{w/o Keyword Aug} &
  \checkmark &
  \checkmark &
  - &
  \checkmark &
  \small{31.43} &
  \small{49.63} &
  \small{31.51} &
   \small{50.56} &
   \small{30.41} &
   \small{52.02} &
   \small{36.34} &
  \small{ 54.57} \\


EPA (\textsc{LLaMA2-7B}) & \checkmark & \checkmark & \checkmark & \checkmark & {31.54}  & \textbf{51.12}  & {31.55} & {52.27} & {30.10} & \textbf{53.55} & {36.22} & {55.49} \\
  
\quad \small{w/o Keyword Aug} &
  \checkmark &
  \checkmark &
  - &
  \checkmark &
  \small{31.12} &
  \small{50.33} &
  \small{31.44} &
  \small{52.07} &
  \small{30.01} &
  \small{53.49} &
  \small{35.70} &
\small{54.90} \\
  
\rowcolor[gray]{0.9} EPA (GPT3.5) & 
  \checkmark &
  \checkmark &
  \checkmark &
  \checkmark &
  \textbf{32.39} &
  \textbf{51.12} &
  \textbf{32.24} &
  \textbf{52.70} &
  \textbf{30.95} &
    53.34 &
  36.61 &
  \textbf{55.87} \\
  
  
\quad \small{w/o Keyword Aug} &
  \checkmark &
  \checkmark &
  - &
  \checkmark &
  \small{31.31} &
  \small{50.67} &
  \small{31.13} &
  \small{52.29} &
  \small{30.06} &
  \small{52.85} &
  \small{35.40} &
  \small{55.80} \\\hline
Olisia \cite{jacqmin2023olisia}         & - & - & - & - & 30.17 & 46.25 & 30.43 & 48.07 & 29.13 & 49.21 & 36.1  & 52.58 \\
\bottomrule
\end{tabular}%
}
\caption{Comparison of various augmentation methods in enhancing the robustness of DST models across different ASR environments. In the feature columns, ``Aud.'' indicates the use of audio-based augmentation, ``Utt-aug.'' denotes utterance-level text augmentation, ``Key-aug.'' refers to keyword-specific augmentation, and ``LLM'' indicates whether large language models were used for error generation. 
All results were averaged over three seeds for better consistency.}
\label{tab:main}
\vspace{-10pt}
\end{table*}

%% file: analysis.tex
\subsection{Robustness Improvement through EPA}

\paragraph{EPA improves robustness. }The results in Table~\ref{tab:main} shows the effectiveness of EPA in  robustness to ASR errors. Remarkably, EPA outperformed existing text-based and audio-based augmentation, showing substantial improvement in JGA and named entity accuracy. It also surpassed the previous best-performing model, Olisia, particularly  in challenging environments.

\paragraph{Effectiveness of keyword-specific error.} In Table~\ref{tab:main}, we present an ablation study to evaluate the effectiveness of keyword-level augmentation. We found that adding keyword-specific ASR errors improved DST performance across all environments and was particularly helpful in enhancing the robustness of named entity accuracy. Additional experiments, including generalization to other backbones and tasks, as well as statistical significance analysis, are provided in Appendix~\ref{app:exp_main}.

\subsection{Comparison with Error Correction Models}
\input{table/table_err_correction}
In Table~\ref{tab:err_correction}, we compare our method with two ASR error correction approaches. For \textsc{Error Correction Model 1}, We fine-tuned a pre-trained on 300 hours of ASR-errored and gold transcription pairs (same as in Section~\ref{sec:method_step1}) in a seq2seq manner, minimizing \( L = - \sum_{i=1}^{N} \log P(g_i \mid e_i) \),
where \(g_i\) and \(e_i\) denote the gold and errored transcriptions, respectively. 
Training used early stopping (patience=3) with batch size 16.
For \textsc{Error Correction Model 2}, we adopted an off-the-shelf\footnote{url{https://huggingface.co/oliverguhr/spelling-correction-english-base}} text correction model to revise typos and transcription errors. Each model corrected the test transcriptions from the weak, noised, and strong ASR systems before DST evaluation. 
As in the table, \textsc{Correction model 1} underperforms the baseline, often altering named entities unnecessarily (e.g., "Grifon" $\rightarrow$ "Bristol"), likely due to limited training diversity. 
\textsc{Correction model 2} yields minor improvements, while our EPA achieves the highest robustness across all ASR conditions.

\subsection{Qualitative Assessment of EPA Method}
\label{sec:quality}
\input{table/table_quality}

\paragraph{Automatic evaluation.} Although Table~\ref{tab:main} confirms EPA's effectiveness, it remains unclear whether the LLM-generated augmentations truly reflect diverse, keyword-focused ASR-style errors. To this end, we perform the quality analysis based on three metrics (Table~\ref{tab:quality}): the unique word increase rate, named entity change rate, and pronunciation similarity with original sentence. The results show that EPA achieves remarkable diversity in unique words (1.81×) and the highest named entity change rate (95.47\%), while maintaining high pronunciation similarity (91.57\%). Notably, keyword-level augmentation plays a key role in enhancing named entity variability, increasing the change rate from 68.81\% to 95.47\%.

\paragraph{Comparison with Authentic ASR Errors.}
\label{app:appendix-edit}
\begin{figure}[h]
  \centering
  \includegraphics[width=.95\columnwidth]{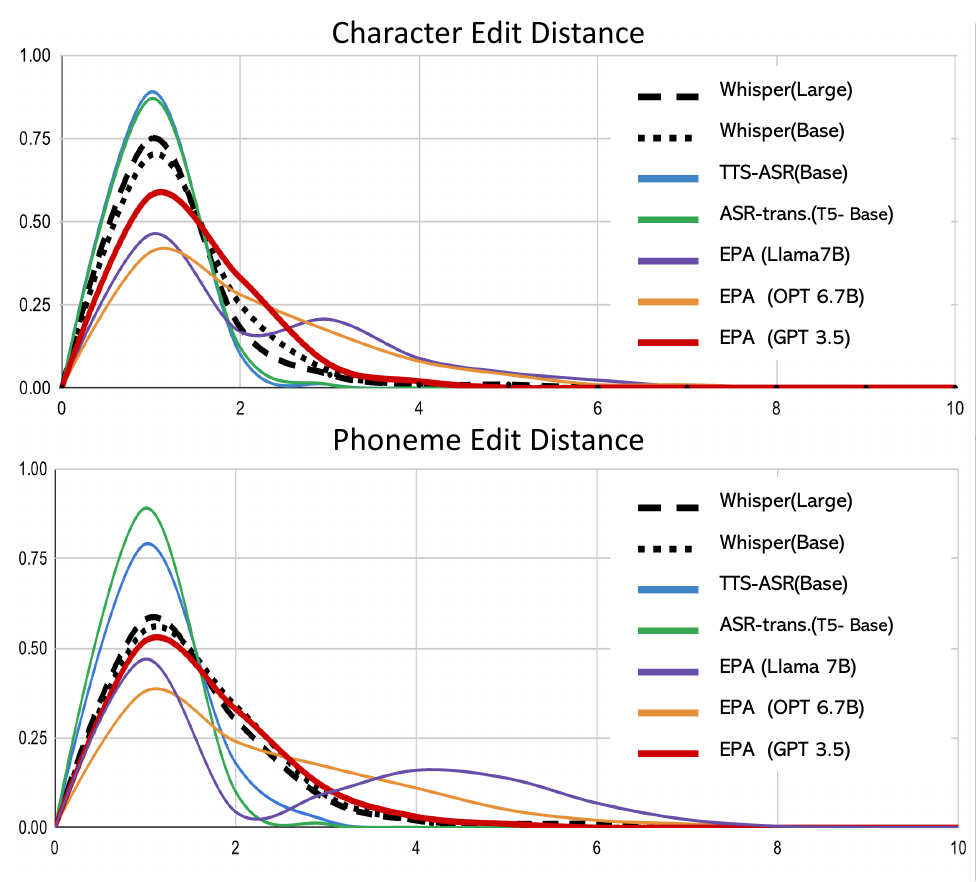}
   \caption{Distribution of edit distance. The x-axis represents edit distances, and the y-axis represents the corresponding ratio.}
   \label{fig:edit}
\end{figure}
\input{table_edit}

In this analysis, we explored the similarity between simulated ASR errors and authentic ASR errors from the perspective of edit distance. Specifically, we examined the distribution of edit distances in simulated data and in errors produced by Whisper large/base models, considering both character- and phoneme-level representations (Figure~\ref{fig:edit}). To quantify the distributional differences, we computed the Jensen-Shannon Divergence (JSD), a symmetric variant of the Kullback-Leibler divergence (Table~\ref{tab:edit}).

Our experiments yielded several noteworthy findings.
Notably, the LLM-simulated errors closely matched the distribution of real ASR errors, particularly at the phoneme level.
This suggests that LLMs can effectively capture pronunciation-level variations and generate realistic ASR-style distortions.
In contrast, errors produced by TTS-based synthesis or ASR-translation models exhibited larger divergence from real ASR patterns, likely due to their tendency to generate more diverse but less phonetically grounded outputs.

\paragraph{Human evaluation.} To further verify the quality of our EPA method, we conducted a human evaluation using 100 sentence pairs, each consisting of an original sentence and its augmented counterpart, with two human evaluators. Participants rated how likely the change resembled an ASR error on a 4-point Likert scale, where 1 indicated "not like an ASR error" and 4 indicated "clearly an ASR error." The average rating was 3.22 with moderate inter-rater agreement   (Gwet’s AC2 \cite{gwet} = 0.590), suggesting that most EPA-generated edits were perceived as realistic ASR errors. Details on the evaluation metric and human evaluation are provided in Appendix~\ref{app:quality}.

\subsection{Error Analysis}

We additionally analyze the impact of keyword augmentation by examining how it influences specific error types in DST predictions. Table~\ref{tab:err} presents the percentage reduction in error rates compared to the baseline. The results demonstrate that EPA is effective in "Wrong" and "Ignore" error types, and keyword augmentation highly contributed to this improvement by decreasing the error rate from 5.29\% to 8.19\%. Interestingly, while keyword augmentation led to substantial reductions in "Wrong" errors, it also caused a slight increase in "Spurious" errors. This may be because the model, after repeatedly seeing phonetic noise around slot values, becomes overly sensitive and starts hallucinating unmentioned slots. A potential mitigation is to introduce an additional loss term for slot presence prediction \cite{trippy,kim2019efficient} , helping the model better distinguish between mentioned and unmentioned slots.
\input{table_err}

%% file: table/table_err_correction.tex
\begin{table}[t]
\centering
\resizebox{\columnwidth}{!}{%
\begin{tabular}{l|cc|cc|cc}
\hline
\multicolumn{1}{c|}{\multirow{2}{*}{Method}} &
  \multicolumn{2}{c|}{Weak-ASR} &
  \multicolumn{2}{c|}{Noised-ASR} &
  \multicolumn{2}{c|}{Strong-ASR} \\\cline{2-7} 
\multicolumn{1}{c|}{} & JGA & N-acc & JGA & N-acc & JGA & N-acc \\ \hline
Baseline & 29.88 & 45.76 & 29.70 & 46.77 & 34.87 & 52.07 \\ 
+ \textsc{Correction model 1} & 23.30 & 41.50 & 23.96 & 42.86 & 26.68 & 46.14 \\ 
+ \textsc{Correction model 2} & 30.14 & 46.79 & 29.82 & 47.68 & 34.62 & 52.00 \\ 
+ EPA (with GPT, ours) & \textbf{32.39} & \textbf{51.12} & \textbf{32.24} & \textbf{52.70} & \textbf{36.61} & \textbf{55.87} \\ \hline
\end{tabular}%
}
\caption{Comparison with ASR error correction models under different ASR conditions. }
\label{tab:err_correction}
\end{table}

%% file: table/table_quality.tex
\begin{table}[!ht]
\centering
\resizebox{0.9\columnwidth}{!}{%
\begin{tabular}{l|rrr}
\hline
Method & \multicolumn{1}{r|}{\makecell{Uniq. \\ Words}} & \multicolumn{1}{r}{\makecell{NE.chg \\ \text{[ \% ]}}}  & \multicolumn{1}{r}{\makecell{Pronoun  \\ Sim.\text{[\%]}}}  \\ \hline
Baseline         & 1                          & -    & -  \\\hline
AEDA \cite{aeda} & \multicolumn{1}{l}{1.00$\times$} & 44.29   & 91.57         \\
EDA \cite{eda}   & \multicolumn{1}{l}{0.86$\times$} & 70.03   & 61.14         \\
BT \cite{BT}     & \multicolumn{1}{l}{1.21$\times$} & 73.46   & 77.17         \\
TTS-ASR       & \multicolumn{1}{l}{1.01$\times$} & 38.84 & \textbf{98.93}  \\
Translating           & \multicolumn{1}{l}{0.84$\times$} & 39.59 & 94.07           \\
\rowcolor[gray]{0.9}  EPA     & \multicolumn{1}{l}{\textbf{1.81}$\times$} & \textbf{95.47} & 91.57  \\
\quad w/o Keyword Err. & \multicolumn{1}{l}{1.57$\times$} & 68.81  & 93.14      \\ 
\hline
\end{tabular}%
}
\caption{Assessment of EPA dataset quality: Unique word increased rate, Named entity changed rate (NE.chg), and pronunciation similarity. }
\label{tab:quality}
\end{table}


%% file: table_edit.tex
\begin{table}[t]
\centering
\resizebox{.9\columnwidth}{!}{%
\begin{tabular}{l|rr|rr}
\toprule
\multirow{2}{*}{Method} & \multicolumn{2}{c|}{Text Dist.($\downarrow$)}                     & \multicolumn{2}{c}{Phoneme Dist.($\downarrow$)}                  \\\cline{2-5}
                        & \multicolumn{1}{c}{ASR-L} & \multicolumn{1}{c|}{ASR-B} & \multicolumn{1}{c}{ASR-L} & \multicolumn{1}{c}{ASR-B} \\\hline
TTS-ASR \small{(Whisper-B)}  & 0.030  & 0.048 & 0.056          & 0.061        \\
TTS-ASR \small{(Whisper-S)} & 0.030  & 0.048 & 0.070           & 0.077          \\
ASR trans. \small{(T5-base)}            & \textbf{0.025} & 0.039 & 0.104          & 0.116          \\ \hline
EPA \small{(Llama2 7B)}    & 0.218 & 0.123 & 0.204          & 0.256          \\
EPA \small{(OPT 6.7B)}      & 0.115 & 0.106 &  0.071  &  0.091    \\
EPA \small{(GPT 3.5)}      & 0.033 & \textbf{0.010}  & \textbf{0.009} & \textbf{0.007} \\ 

\toprule
\end{tabular}%
}
\caption{Distribution distance (JSD) between Whisper Large/Base model and simulation dataset. (ASR‑L=Whisper Large, ASR‑B=Whisper Base). }
\label{tab:edit}
\end{table}

%% file: table_err.tex
\begin{table}[t]
\centering
\resizebox{0.95\columnwidth}{!}{%
\footnotesize{
\begin{tabular}{lccc}
\hline
\multicolumn{1}{l|}{\multirow{2}{*}{Method}} &
  \multicolumn{3}{c}{Error Type} \\ \cline{2-4} 
\multicolumn{1}{l|}{} &
  \multicolumn{1}{l|}{Wrong} &
  \multicolumn{1}{l|}{Ignore} &
  \multicolumn{1}{l}{Spurious} \\ \hline

  \multicolumn{4}{c}{Noised Audio} \\ \hline
\multicolumn{1}{l|}{Baseline} &
  \multicolumn{1}{c|}{\begin{tabular}[c]{@{}c@{}}{\color{blue}$\triangledown$}0\%\\ \scriptsize{(6237)}\end{tabular}} &
  \multicolumn{1}{c|}{\begin{tabular}[c]{@{}c@{}}{\color{blue}$\triangledown$}0\%\\ \scriptsize{(3654)}\end{tabular}} &
  \begin{tabular}[c]{@{}c@{}}{\color{blue}$\triangledown$}0\%\\ \scriptsize{(2027)}\end{tabular} \\ \hline
\multicolumn{1}{l|}{EPA w/o Key-aug} &
  \multicolumn{1}{c|}{\begin{tabular}[c]{@{}c@{}}{\color{blue}$\triangledown$}5.29\%\\ \scriptsize{(5907)}\end{tabular}} &
  \multicolumn{1}{c|}{\begin{tabular}[c]{@{}c@{}}{\color{blue}$\triangledown$}3.72\%\\ \scriptsize{(3518)}\end{tabular}} &
  \begin{tabular}[c]{@{}c@{}}{\color{blue}$\triangledown$}\textbf{6.31}\%\\ \scriptsize{(1899)}\end{tabular} \\ \hline
\multicolumn{1}{l|}{EPA} &
  \multicolumn{1}{c|}{\begin{tabular}[c]{@{}c@{}}{\color{blue}$\triangledown$}\textbf{8.19}\%\\ \scriptsize{(5726)}\end{tabular}} &
  \multicolumn{1}{c|}{\begin{tabular}[c]{@{}c@{}}{\color{blue}$\triangledown$}\textbf{7.25}\%\\ \scriptsize{(3389)}\end{tabular}} &
  \begin{tabular}[c]{@{}c@{}}{\color{blue}$\triangledown$}1.33\%\\ \scriptsize{(2000)}\end{tabular} \\ \hline

\end{tabular}
}
}
\caption{Ablation study with error analysis. \textbf{Wrong} indicates the model predicts incorrect values, \textbf{Ignore} refers to ignored mentioned slots, and \textbf{Spurious} denotes predicting values for unmentioned slots.  Actual error numbers are in parentheses.}
\label{tab:err}
\vspace{-14pt}
\end{table}


%% file: appendix.tex




\section{Details of the EPA Method}

\subsection{Prompt Used for EPA}
\label{app:prompt}
The prompts used in Step 1 and Step 2 are provided below.

\input{prompt/prompt1}

\input{prompt/prompt2}

\subsection{Detailed Keyword Highlighting Strategy for EPA}

\input{table/hl_example.tex}

To explicitly introduce keyword-specific ASR errors, we first identify dialogue state values from the training corpus and match them against the user utterance ($\dot{u}$). Matched values are then automatically wrapped with \texttt{<hl>} tags based on slot annotations (e.g., DST slot labels or NER tags), as shown in Table~\ref{tab:hl_example}. These highlighted utterances are passed to the LLM, which is instructed to perturb the text within the \texttt{<hl>} tags while preserving the rest. This keyword highlighting strategy is task-agnostic and can be easily applied to other keyword-sensitive tasks such as Named Entity Recognition (NER) or Spoken Language Understanding (SLU), where certain slot values or entities are critical for downstream prediction.

\input{table/retreived_examples}

\subsection{Retrieved In-Context Example}
\label{app:EPA_retreive}
In Section~\ref{sec:method_step1}, we retrieved in-context examples based on phoneme-level similarity. Table~\ref{tab:retreive_example}, we present several representative examples to illustrate this retrieval process, showing how phonetically similar phrases (highlighted in color) are matched between the target and retrieved utterances. This demonstrates that the retrieval mechanism effectively captures pronunciation-level patterns relevant to ASR-style errors.

\subsection{Additional Examples of ASR-style Errors}
\input{table/table_asr_examples}

\label{app:asr-examples}

Table~\ref{tab:table_asr_examples} presents additional examples of ASR-style errors generated by our EPA method, including both general and keyword-specific transformations.

\section{Experimental Setup}
\label{app:exp_setting}
\subsection{Details of the ASR Environment}
\begin{itemize}
    \item Low-acc ASR environment: Whisper-base model (74M)\cite{whisper} is used for transcription. WER on LibriSpeech.test-clean is 0.05.
    \item Noisy audio environment: Incorporated authentic cafe and traffic noise from \url{https://freesound.org/} with a 10 to 20 Signal-to-Noise Ratio (SNR) and transcribed it using the Whisper large model.
    \item Paraphrased environment: When recording the audio, the text was paraphrased to resemble more natural, real-life spoken language\cite{dstc11}.
    \item High-acc ASR environment: Whisper-large model (1550M) is used for transcription. WER on LibriSpeech.test-clean is 0.027.
\end{itemize}

\subsection{Comparison Methods} 
\label{sec:appendix-experiment-compared-methods}
\begin{itemize}
    \item AEDA \cite{aeda}: We randomly inserted punctuation marks, effectively maintaining the original word order.
    \item EDA \cite{eda}: We augmented data by applying edit-based technique that implements four rule-based modifications—synonym replacement, random insertion, swapping, and deletion.
    \item Back Translation \cite{BT}: We translated original texts to error texts and then back to the original texts for generating syntactic variations during the process. We use English to German \footnote{\url{facebook/wmt19-en-de}} and German to English\footnote{\url{facebook/wmt19-de-en}} models as translator.

    \item TTS-ASR : We used Tacotron2 \cite{taco} for the TTS model to synthesize the audio and use Whisper-base \cite{whisper} as an ASR model to simulate the ASR errors.

    \item ASR translation: We employed a sequence-to-sequence structure to translate clean text into ASR-errored text. Our training set comprised 300 hours of paired clean and ASR-errored text. We fine-tuned the model based on the T5-base architecture \cite{t5}, using the loss function defined in equation~\ref{eq:label_loss_translation}. The loss function is as follows:
    \begin{equation}\label{eq:label_loss_translation}
      L = -\sum_{i=1}^{I} \log P(e_i|g_i).
    \end{equation}

\end{itemize}

\subsection{Training Details}
\label{sec:app-training}
 In training models, we used T5-base \cite{t5} as the backbone model and instructed the model to generate the $B_t$ by given $D_t$ in sequence to sequence manner, as in \cite{pptod} and the loss function is 
    \begin{equation}\label{eq:label_loss}
      L = -\sum_{t=1}^{T} \log P(B_t|\text{Inst}, D_t).
    \end{equation}
We set the learning rate as 4e-5 and used the AdamW \cite{loshchilov2017decoupled} optimizer. One GeForce RTX 3090 is used for training and the batch size is 16. Trained until reaching the max patient, which is 3.

\section{Further Experiments}
\label{app:exp_main}

\subsection{Statistical Significance Analysis}
\input{table/table_static}
To assess the reliability of our results, we conducted paired \textit{t}-tests between each method and the Baseline to determine whether the observed performance improvements are statistically significant. We report 95\% confidence intervals to reflect performance variability. Statistical significance is denoted using asterisks: $\ast$ for $p < 0.05$, $\ast\ast$ for $p < 0.01$, and $\ast\ast\ast$ for $p < 0.001$.

As shown in Table~\ref{tab:significance}, EPA (GPT-3.5) achieves statistically significant gains in nearly all evaluation settings, particularly under low-accuracy and noised ASR conditions. These results confirm that the improvements brought by our method are both consistent and statistically reliable.

\subsection{Baseline Performance Comparison with Clean Text }
For comparison, we report the baseline performance on an error-free, clean test dataset. Please note that DSTC11 \cite{dstc11} does not provide a text script for the test dataset, so we are manually cleaning 50 dialogues to ensure they are error-free. In the experiment, the baseline model achieved a JGA score of 45.2 \% and an N-ACC score of 86.5 \% in an ASR error-free environment. Compared to the JGA, which is 34.8 \%, and N-ACC, which is 52.07 \%, in the ASR-errored environment (High-acc ASR model environment), this discrepancy highlights the significant impact of ASR errors on performance degradation.

\subsection{Effect of Keyword Augmentation}
\input{table/table_keyword_ablation}
To further validate the contribution of each component in our method, we conducted an ablation study separating Step 1 (overall error augmentation) and Step 2 (keyword-specific error augmentation), using GPT-3.5 as the generator. Both augmentation strategies individually improved performance over the baseline. 
Step~1 yielded consistent gains across all ASR settings, while Step~2 notably enhanced slot-value accuracy. When combined (EPA, ours), the model achieved the highest robustness in all evaluation conditions, confirming that the two strategies are complementary.

\subsection{Experiments with a Different Baseline}
\input{table/table_gpt2}

In the main experiments, we use T5-base as the backbone model. To assess the generalizability of our approach, we additionally conduct experiments using a GPT-2~\cite{gpt2} model, as shown in Table~\ref{tab:gpt2}. The results show a consistent trend with those of T5-base, demonstrating that our method is effective across different backbone architectures.

\subsection{Generalizability to Other Tasks}
\label{app:generalizability}
\input{table/table_NER}
\input{table/table_SLU}
To evaluate the generalizability of our approach beyond the DST domain, we extended our experiments to two additional spoken language understanding tasks: Named Entity Recognition (NER) and Spoken Language Understanding (SLU). We applied our EPA methodology under three ASR conditions—low-accuracy ASR, noised audio, and high-accuracy ASR—using the same experimental setup as in the DSTC11 evaluation. We used asapp/slue dataset for NER task\cite{asapp}, and  SLURP dataset \cite{slurp} for SLU task.

The results, shown in Table~\ref{tab:ner_results} and Table~\ref{tab:slu_results}, demonstrate that our method consistently improves performance across all ASR conditions for both NER and SLU tasks. Notably, the gains are especially prominent under low-accuracy and noisy conditions, confirming that our approach is broadly applicable to other tasks.

\subsection{Additional Fine-grained Metrics}
\label{appendix:finegrained}
\input{table/table_fine}
To supplement the main results focusing on JGA and N-Acc, we report additional fine-grained metrics—Precision, Recall, F1, and Slot Accuracy—under two ASR corruption settings: Low-accuracy ASR and Noised ASR. These metrics provide a more comprehensive view of model behavior in diverse error conditions in table~\ref{tab:fine_1} and ~\ref{tab:fine_2}.

\subsection{Results with Different Random Seeds}
\input{table/table_seeds}

Table~\ref{tab:seeds} reports the results of our main experiments (Table~\ref{tab:main}) repeated with three different random seeds, demonstrating the consistency of the observed trends.

\section{Details of Quality Evaluation}
\label{app:quality}
\subsection{About Metric}

As described in Section~\ref{sec:quality}, we use a phonetic similarity metric to evaluate pronunciation-level consistency between the original and augmented text. Specifically, we compute the normalized phoneme edit distance, which quantifies the minimal number of phoneme-level operations required to transform one utterance into another. A higher score indicates greater phonetic similarity.
We used the \texttt{eng-to-ipa} library\footnote{\url{https://pypi.org/project/eng-to-ipa/}} for phoneme conversion in our implementation, as shown in the code snippet below.

\begin{lstlisting}[style=pythonstyle]
def phonetic_similarity(original_text, augmented_text):
    
    original_ipa  = to_phoneme(original_text)
    augmented_ipa = to_phoneme(augmented_text)
    
    edit_distance = nltk.edit_distance(original_ipa, augmented_ipa)
    
    # Normalize the edit distance
    max_length = max(len(original_ipa), len(augmented_ipa))
    normalized_distance = float(edit_distance) / float(max_length)
    
    # Convert to similarity score
    similarity_score = 1 - normalized_distance
    return similarity_score
\end{lstlisting}

\subsection{Human Evaluation Details}

To assess the plausibility of the generated ASR-style errors, we conducted a human evaluation involving two graduate students. Each participant was asked to rate whether a given sentence transformation could plausibly be attributed to an ASR error, using a 4-point Likert scale:

\begin{itemize}
    \item \textbf{1 – Not at all}: The change is unlikely to be due to an ASR error. It appears to stem from other factors such as meaning alteration or stylistic variation.
    \item \textbf{2 – Unlikely}: The transformation is probably not caused by an ASR error.
    \item \textbf{3 – Somewhat likely}: The transformation may plausibly be caused by an ASR error.
    \item \textbf{4 – Very likely}: The transformation clearly appears to result from an ASR error.
\end{itemize}

Each sentence pair (original and transformed) was rated independently by both annotators. Inter-rater agreement and average scores are reported in Section~\ref{sec:quality}. The distribution of Likert scores for each annotator is as follows : Annotator 1 assigned 5\% of scores as 1, 7\% as 2, 17\% as 3, and 71\% as 4. Annotator 2 assigned 3\% of scores as 1, 24\% as 2, 53\% as 3, and 20\% as 4.

\normalsize

%% file: prompt/prompt1.tex
\begin{prompt}[colback=black!0!white, colframe=black!98!black]
{Step 1 Prompt}{\textcolor{myBlue}{Generate ASR error augmented text with similar pronunciation but different words based on the given gold text examples.}\\
Apply character and word substitutions, additions, or deletions while maintaining the overall pronunciation and context. \\
Error rate should be high 
\par            
\noindent\dotfill\par 
\textbf{Example 1}\\
Original: they have a single naupliar eye  \\
ASR-errored: they have a single nor pure eye \\

\textbf{Example 2}\\
Original: i must have saint louis then huzza  \\
ASR-errored: i must have st louis then hazard \\

\textbf{Example 3}\\
Original: i wonder uncle did not have her come  \\
ASR-errored: i wonder uncle did not have a problem 
\par            
\noindent\dotfill\par 
\textcolor{myBlue}{Now, following the above examples, generate an ASR-errored version of the following sentence:\\}
Original: \textcolor{myOrange}{[Target utterance]} \\
ASR-errored: 
}
\end{prompt}

%% file: prompt/prompt2.tex
\begin{prompt}[colback=black!0!white, colframe=black!98!black]
{Step 2 Prompt}{\textcolor{myBlue}{Change the key words in <hl> tag, to having a ASR error.
ASR error has similar pronounciation with the correct word, but different charater. }\\
Here is some example.
\par            
\noindent\dotfill\par 
\textbf{Example 1}\\
Original: I want to buy a book about <hl>luwombo best</hl> restaurant. \\
Keywords : luwombo best \\
Result: I want to buy a book about luwambo vest restaurant.  \\\\
\textbf{Example 2}\\
Original: hi, i'm looking for a bus that is depart from <hl>eliot<hl/> and arriving to <hl>holiday inn williamsport<hl/>? \\
Keywords : eliot, holiday inn williamsport \\
Result:  hi, i'm looking for a bus that is depart from Ellyot and arriving to holliday inn william's port \\\\
\textbf{Example 3}\\
Original: the <hl>chabuton ramen<hl/> is a restaurant on the east. \\
Keywords : chabuton ramen \\
Result: the shabuton raymond is a restaurant on the east.\\ 
\par            
\noindent\dotfill\par 
\textcolor{myBlue}{Now, following the above examples, generate an ASR-errored version of the following sentence:\\}
Original: \textcolor{myOrange}{[Target utterance with <hl> tag]} \\
ASR-errored: 
}
\end{prompt}


%% file: table/hl_example.tex

\begin{table}[h!]
\small
\resizebox{\columnwidth}{!}{
\begin{tabular}{p{0.2\columnwidth} p{0.8\columnwidth}}
\toprule
\multicolumn{2}{l}{\textbf{Example of adding <hl> tag}} \\
\midrule
Original & Hi, I need to go to Green Day hotel, then book a table at the Grill House. \\
Dialogue State & hotel-name: \textcolor{myGreen}{Green Day}, restaurant-name:\textcolor{myBlue}{Grill House} \\
\cdashline{1-2}
With \texttt{<hl>} tags & Hi, I need to go to \texttt{<hl>}\textcolor{myGreen}{Green Day}\texttt{</hl>} hotel, then book a table at the \texttt{<hl>}\textcolor{myBlue}{Grill House}\texttt{</hl>}. \\
\bottomrule
\end{tabular}
}
\caption{Example of keyword highlighting using \texttt{<hl>} tags based on dialogue state annotations.}
\label{tab:hl_example}
\end{table}

%% file: table/retreived_examples.tex
\begin{table*}[h!]
\small
\resizebox{\textwidth}{!}{
\begin{tabular}{p{0.2\textwidth}p{0.8\textwidth}}

\toprule
\multicolumn{2}{l}{Example 1}     \\ \toprule
                                            
Target & \textcolor{myOrange}{can you tell} me the address to the \textcolor{myRed}{police station} \textcolor{myGreen}{in point pleasant?}  \\ \cdashline{1-2}

Retrieved 1 &  \textcolor{myRed}{frayser station} was not the depot \textcolor{myGreen}{on the point}      \\
\quad \small{+ ASR} & freya station was not the deep watch on the point   \\

Retrieved 2&  \textcolor{myOrange}{can you get me} the maldeamores saga    \\
\quad \small{+ ASR} & can you get me the melamorphos     \\
Retrieved 3               &  \textcolor{myOrange}{cannot you tell} her whom i am eh joseph\\
\quad \small{+ ASR} & cannot you tell her whom i am    \\

\toprule
\multicolumn{2}{l}{Example 2}     \\ \toprule
                                            
Target & no, i just \textcolor{myBlue}{need} to make \textcolor{myGreen}{sure} it's cheap. oh, and i need parking. \\ \cdashline{1-2}

Retrieved 1 &  i \textcolor{myBlue}{need} fifty ten foot long \textcolor{myGreen}{segments} of wire    \\
\quad \small{+ ASR} & i need fifty ten foot long signals of my life   \\

Retrieved 2&  a drive with a different encoding mechanism would \textcolor{myBlue}{need} different patterns  \\
\quad \small{+ ASR} & and drive was a different building recognition would need different patterns    \\
Retrieved 3               &  to  \textcolor{myBlue}{reach} to calcutta you  \textcolor{myBlue}{need} less time to reach dhaka   \\
\quad \small{+ ASR} & to reach tukaukara you need last time to reach daka    \\

\toprule
\multicolumn{2}{l}{Example 3}     \\ \toprule
                                            
Target & i'm \textcolor{myBlue}{open to} any \textcolor{myOrange}{kind of} food. i'm looking for something in the \textcolor{myGreen}{centre} and on the expensive side. \\ \cdashline{1-2}

Retrieved 1 &  \textcolor{myOrange}{kokai} means \textcolor{myBlue}{open to} the public or laid \textcolor{myBlue}{open}      \\
\quad \small{+ ASR} & cook eye means open to the public all laid open  \\

Retrieved 2&   the town of beauharnois was the major \textcolor{myGreen}{centre}    \\
\quad \small{+ ASR} & the town of bo hanwa was the major center    \\
Retrieved 3               &  the gate is \textcolor{myBlue}{open} at eleven    \\
\quad \small{+ ASR} & the gate is open at 11   \\
\bottomrule
\end{tabular}
}
\caption{In-context examples retrieved based on phoneme-level similarity.
For each target utterance (top row), we retrieve three ($g$, $e$) example pairs from the database using phonetic similarity between the target and $g$.  Colored segments highlight phonetically similar phrases between the target and retrieved examples.}
\label{tab:retreive_example}
\end{table*}

%% file: table/table_asr_examples.tex
\begin{table*}[h!]
\small
\resizebox{\textwidth}{!}{
\begin{tabular}{p{0.5\textwidth}p{0.5\textwidth}}

\toprule
\textbf{Original} & \textbf{Augmented} \\\hline

- no, i just need to make sure it's \textcolor{myOrange}{cheap}. oh, and i need parking  
& - no, i just need to make sure it's \textcolor{myOrange}{sheep}. oh, and i need parking. \\

- i am departing from \textcolor{myBlue}{marion}  - 
& - i am departing from \textcolor{myBlue}{maryland} \\

- no, i don't need  \textcolor{myOrange}{anything} else right now. thank you for your  \textcolor{myOrange}{assistance}. good  \textcolor{myOrange}{bye}.  
& - no, i don't need \textcolor{myOrange}{anyone} else right now. thank you for your \textcolor{myOrange}{persistence}. good \textcolor{myOrange}{buy}. \\

- i would like to go to \textcolor{myBlue}{sandy} please.  
& - i would like to go to \textcolor{myBlue}{cindy} please. \\

- i would like to keep it in the \textcolor{myBlue}{moderate} range, please.  
& - i would like to keep it in the \textcolor{myBlue}{mod rain}, please \\

- could i get the \textcolor{myOrange}{address} for it? i would also like an \textcolor{myBlue}{expensive} place to eat around it.  
& - could i get the \textcolor{myOrange}{actress} for it? i would also like an \textcolor{myBlue}{extensive} place to eat around it. \\

- i need to take a \textcolor{myBlue}{train} out of garrett, i will be leaving town on wednesday
& - i need to make a \textcolor{myBlue}{plane} out of garrett, i will be \textcolor{myOrange}{weaving} town on wednesday. \\

- do you have any \textcolor{myBlue}{indian} restaurants in the \textcolor{myBlue}{south} in a different \textcolor{myOrange}{price} range?  
& - do you have any \textcolor{myBlue}{indonesian} restaurants in the \textcolor{myBlue}{south} in a different \textcolor{myOrange}{prize} range \\

- nope, same \textcolor{myOrange}{people}.  
&-  nope, same \textcolor{myOrange}{pupil}. \\

- i'm looking for a \textcolor{myBlue}{college} type \textcolor{myBlue}{attraction}  
& - i'm looking for a \textcolor{myBlue}{knowledge}-type \textcolor{myBlue}{action}. \\

- yes, please \textcolor{myOrange}{book} me a \textcolor{myOrange}{room} for friday
& - yes, please \textcolor{myOrange}{cook} me a \textcolor{myOrange}{broom} for friday \\

- yes, could you please email or \textcolor{myOrange}{fax} me the \textcolor{myOrange}{fare} amount, as well as the reference number?  
& - yes, could you please email or \textcolor{myOrange}{text} me the \textcolor{myOrange}{fair} amount, as well as thereference code? \\

- \textcolor{myBlue}{ois el shaddai} a \textcolor{myOrange}{guest house} or \textcolor{myBlue}{hotel}?  
& \textcolor{myBlue}{iz let shadai} a \textcolor{myOrange}{gest house} or \textcolor{myBlue}{motel}? \\

- great! i also need a \textcolor{myBlue}{train} from \textcolor{myBlue}{mount pleasant} to \textcolor{myBlue}{sabattus}, please.  
& - great! i also need a \textcolor{myBlue}{strain} from \textcolor{myBlue}{mount pleasant} to \textcolor{myBlue}{suspicious}, please \\

- yes, can you help me find a \textcolor{myBlue}{train} that can take me from \textcolor{myBlue}{lovelock} to \textcolor{myBlue}{abbot}?  
& - yes, can you help me find a \textcolor{myBlue}{plane} that can take me from \textcolor{myBlue}{love lock} to \textcolor{myBlue}{rabbit}? \\

\bottomrule
\end{tabular}
}
\caption{
Examples of augmented utterances generated by injecting phoneme-level ASR-style errors. For each original utterance (left), the corresponding augmented version (right) includes substitutions that mimic realistic ASR recognition mistakes.  
\textcolor{myBlue}{Blue-colored phrases} indicate changes in keywords that are used as slot values in DST, while \textcolor{myOrange}{orange-colored phrases} represent overall ASR-style errors.
}

\label{tab:table_asr_examples}
\end{table*}

%% file: table/table_static.tex
\begin{table}[t]
\centering
\resizebox{\columnwidth}{!}{%
\begin{tabular}{l|cc|cc|cc|cc}
\hline
\multicolumn{1}{c|}{\multirow{2}{*}{Method}} &
  \multicolumn{2}{c|}{Low-acc ASR} &
  \multicolumn{2}{c|}{Noised Aud.} &
  \multicolumn{2}{c|}{Paraphrased} &
  \multicolumn{2}{c}{High-acc ASR} \\ \cline{2-9} 
\multicolumn{1}{c|}{} & JGA   & N-acc & JGA   & N-acc & JGA   & N-acc & JGA   & N-acc \\ \hline
Baseline              & --    & --    & --    & --    & --    & --    & --    & --    \\ \hline
AEDA                  & ns    & ns    & ns    & ns    & ns    & ns    & ns    & ns    \\
EDA                   & *     & *     & **    & **    & ns    & **    & **    & **    \\
BT                    & **    & **    & **    & **    & *     & **    & **    & *     \\
TTS-ASR               & ns    & ns    & ns    & ns    & ns    & ns    & ns    & ns    \\
ASR trans.                   & ns    & *     & ns    & ns    & ns    & *     & *     & *     \\
EPA (GPT-3.5)         & ***   & **    & ***   & ***   & **    & **    & *     & **    \\
\hline
\end{tabular}%
}
\caption{Statistical significance results compared to the Baseline using paired \textit{t}-tests across three random seeds. Stars indicate significance levels: * for $p<0.05$, ** for $p<0.01$, *** for $p<0.001$, and \textit{ns} for non-significant differences.}
\label{tab:significance}
\end{table}

%% file: table/table_keyword_ablation.tex
\begin{table}[t]
\centering
\resizebox{\columnwidth}{!}{%
\begin{tabular}{l|cc|cc|cc}
\hline
\multirow{2}{*}{Method} &
\multicolumn{2}{c|}{Weak-ASR } &
\multicolumn{2}{c|}{Noised Audio} &
\multicolumn{2}{c}{Strong  ASR} \\ \cline{2-7}
 & JGA & N-acc & JGA & N-acc & JGA & N-acc \\ \hline
Baseline & 29.88 & 45.76 & 29.70 & 46.77 & 34.87 & 52.07 \\
+ Overall Error (Step 1) & 31.31 & 50.67 & 31.13 & 52.29 & 35.40 & 55.80 \\
+ Keyword Error (Step 2) & 31.66 & 49.88 & 31.61 & 51.02 & 36.12 & 54.23 \\
EPA (Step1+2) & \textbf{32.39} & \textbf{51.12} & \textbf{32.24} & \textbf{52.70} & \textbf{36.61} & \textbf{55.87} \\ \hline
\end{tabular}%
}
\caption{Ablation on overall and keyword-specific augmentation steps. Combining both (EPA) yields the highest robustness across ASR conditions.}
\label{tab:keyword_ablation}
\end{table}

%% file: table/table_gpt2.tex
\begin{table}[t]
\centering
\resizebox{\columnwidth}{!}{%
\begin{tabular}{l|cc|cc|cc}
\hline
\multicolumn{1}{c|}{\multirow{2}{*}{Method}} &
  \multicolumn{2}{c|}{Low-acc ASR} &
  \multicolumn{2}{c|}{Noised Aud.} &
  \multicolumn{2}{c|}{Paraphrased} \\\cline{2-7} 
\multicolumn{1}{c|}{} & JGA   & N-acc & JGA   & N-acc & JGA   & N-acc  \\ \hline
Baseline              &29.9 & 45.82 & 27.25 & 41.81 & 25.81 & 44.51 \\ 
\quad + EPA             &30.63 & 48.54 & 27.5 & 46.52 & 27.65 & 47.96 \\ \hline
\end{tabular}%
}
\caption{Experiment with GPT-2 model as baseline.}
\label{tab:gpt2}
\end{table}


%% file: table/table_NER.tex
\begin{table}[t]
\centering
\resizebox{\columnwidth}{!}{%
\begin{tabular}{lccc}
\hline
\textbf{Method} & \textbf{Low-acc ASR} & \textbf{Noised Audio} & \textbf{High-acc ASR} \\
\hline
Baseline             & 56.29 & 60.50 & 60.02 \\
+ OPT 6.7B           & 59.64 & 62.84 & 62.47 \\
+ LLaMA 7B           & 58.29 & 60.53 & 60.46 \\
+ GPT-3.5 (125B)     & 59.39 & 62.62 & 62.16 \\
\hline
\end{tabular}
}
\caption{NER results on the ASAPP/SLUE dataset under different ASR conditions. We reported the F1 score.}
\label{tab:ner_results}
\end{table}

%% file: table/table_SLU.tex
\begin{table}[t]
\centering
\resizebox{\columnwidth}{!}{%
\begin{tabular}{lccc}
\hline
\textbf{Method} & \textbf{Low-acc ASR} & \textbf{Noised Audio} & \textbf{High-acc ASR} \\
\hline
Baseline             & 55.25 & 63.23 & 64.32 \\
+ OPT 6.7B           & 56.97 & 64.94 & 66.10 \\
+ LLaMA 7B           & 57.24 & 64.26 & 65.56 \\
+ GPT-3.5 (125B)     & 58.78 & 66.16 & 67.49 \\
\hline
\end{tabular}
}
\caption{SLU results on the SLURP dataset under different ASR conditions. We reported F1 score.}
\label{tab:slu_results}
\end{table}

%% file: table/table_fine.tex
\begin{table}[h]
\centering
\resizebox{\columnwidth}{!}{%
\begin{tabular}{lcccc}
\hline
\textbf{Method} & \textbf{Precision} & \textbf{Recall} & \textbf{F1} & \textbf{Slot Accuracy} \\
\hline
Baseline         & 50.5 & 50.1 & 50.3 & 92.8 \\
+ TTS-ASR          & 49.5 & 49.6 & 49.5 & 93.4 \\
+ ASR-Translation  & 51.8 & 51.7 & 51.7 & 93.1 \\
+ EPA (GPT3.5)     & 55.2 & 54.8 & 55.0 & 93.7 \\
\hline
\end{tabular}
}
\caption{Fine-grained DST metrics under Low-accuracy ASR setting.}
\label{tab:fine_1}
\end{table}

\begin{table}[h]
\centering
\resizebox{\columnwidth}{!}{%
\begin{tabular}{lcccc}
\hline
\textbf{Method} & \textbf{Precision} & \textbf{Recall} & \textbf{F1} & \textbf{Slot Accuracy} \\
\hline
Baseline         & 52.0 & 51.3 & 51.6 & 92.5 \\
+ TTS-ASR          & 53.7 & 53.1 & 53.4 & 92.8 \\
+ ASR-Translation  & 53.8 & 53.0 & 53.4 & 92.5 \\
+ EPA (GPT3.5)     & 56.1 & 55.3 & 55.7 & 93.2 \\
\hline
\end{tabular}
}
\caption{Fine-grained DST metrics under Noised ASR setting.}
\label{tab:fine_2}
\end{table}

%% file: table/table_seeds.tex
\begin{table}[t]
\centering
\resizebox{\columnwidth}{!}{%
\begin{tabular}{l|cc|cc|cc|cc}
\hline
\multicolumn{1}{c|}{\multirow{2}{*}{Method}} &
  \multicolumn{2}{c|}{Low-acc ASR} &
  \multicolumn{2}{c|}{Noised Aud.} &
  \multicolumn{2}{c|}{Paraphrased} &
  \multicolumn{2}{c}{High-acc ASR} \\ \cline{2-9} 
\multicolumn{1}{c|}{} & JGA   & N-acc & JGA   & N-acc & JGA   & N-acc & JGA   & N-acc \\ \hline
Baseline              & 30.05 & 46.48 & 29.80 & 47.49 & 29.08 & 48.85 & 34.73 & 52.31 \\ \hline
AEDA                  & 29.99 & 46.53 & 29.80 & 47.60 & 28.95 & 49.02 & 34.94 & 52.40 \\
EDA                   & 29.16 & 47.67 & 28.93 & 49.50 & 28.12 & 50.10 & 33.74 & 53.66 \\
BT                    & 31.54 & 49.21 & 31.43 & 51.25 & 29.90 & 51.60 & 36.25 & 54.85 \\
TTS-ASR               & 30.09 & 46.16 & 30.32 & 47.74 & 29.26 & 49.31 & 35.28 & 51.95 \\
ASR trans.                        & 29.98 & 47.70 & 29.95 & 48.49 & 29.72 & 50.43 & 34.82 & 53.35 \\
EPA (GPT3.5)          & 32.56 & 51.62 & 32.27 & 53.48 & 31.10 & 53.91 & 36.65 & 56.14 \\ \hline \hline
Baseline              & 29.82 & 45.63 & 29.75 & 46.20 & 28.55 & 48.75 & 34.83 & 51.44 \\ \hline
AEDA                  & 29.77 & 46.29 & 29.69 & 47.40 & 29.26 & 48.93 & 34.93 & 52.26 \\
EDA                   & 29.27 & 47.48 & 28.57 & 49.41 & 28.20 & 49.99 & 33.59 & 53.68 \\
BT                    & 31.66 & 49.16 & 31.05 & 50.81 & 29.88 & 51.75 & 36.19 & 54.73 \\
TTS-ASR               & 29.75 & 45.99 & 29.80 & 46.87 & 29.10 & 48.12 & 34.97 & 51.49 \\
ASR trans.                        & 30.86 & 47.47 & 30.52 & 48.53 & 29.63 & 50.11 & 35.73 & 53.59 \\
EPA (GPT3.5)          & 32.33 & 50.74 & 32.41 & 52.28 & 31.02 & 53.15 & 36.75 & 55.82 \\ \hline\hline
Baseline              & 29.77 & 45.18 & 29.56 & 46.61 & 29.12 & 48.76 & 35.05 & 52.47 \\ \hline
AEDA                  & 29.94 & 46.55 & 29.72 & 47.45 & 29.16 & 48.62 & 34.94 & 52.29 \\
EDA                   & 29.22 & 47.79 & 28.59 & 49.61 & 27.92 & 49.88 & 33.71 & 53.99 \\
BT                    & 31.86 & 49.13 & 31.31 & 50.87 & 29.91 & 51.85 & 36.38 & 54.85 \\
TTS-ASR               & 29.98 & 46.86 & 29.84 & 47.51 & 28.88 & 49.20 & 34.97 & 52.66 \\
ASR trans.                        & 30.37 & 47.78 & 29.94 & 48.32 & 29.26 & 50.60 & 35.20 & 54.05 \\
EPA (GPT3.5)          & 32.27 & 51.01 & 32.04 & 52.33 & 30.72 & 52.96 & 36.42 & 55.66 \\ \hline
\end{tabular}%
}
\caption{Experiment result with different seeds.}
\label{tab:seeds}
\end{table}

%% file: main.bbl
\begin{thebibliography}{37}
\providecommand{\natexlab}[1]{#1}

\bibitem[{Ardila et~al.(2020)Ardila, Branson, Davis, Kohler, Meyer, Henretty, Morais, Saunders, Tyers, and Weber}]{ardila-etal-2020-common}
Rosana Ardila, Megan Branson, Kelly Davis, Michael Kohler, Josh Meyer, Michael Henretty, Reuben Morais, Lindsay Saunders, Francis Tyers, and Gregor Weber. 2020.
\newblock \href {https://aclanthology.org/2020.lrec-1.520} {Common voice: A massively-multilingual speech corpus}.
\newblock In \emph{Proceedings of the Twelfth Language Resources and Evaluation Conference}, pages 4218--4222, Marseille, France. European Language Resources Association.

\bibitem[{Bastianelli et~al.(2020)Bastianelli, Vanzo, Swietojanski, and Rieser}]{slurp}
Emanuele Bastianelli, Andrea Vanzo, Pawel Swietojanski, and Verena Rieser. 2020.
\newblock \href {https://doi.org/10.18653/v1/2020.emnlp-main.588} {{SLURP}: A spoken language understanding resource package}.
\newblock In \emph{Proceedings of the 2020 Conference on Empirical Methods in Natural Language Processing (EMNLP)}, pages 7252--7262, Online. Association for Computational Linguistics.

\bibitem[{Brown et~al.(2020)Brown, Mann, Ryder, Subbiah, Kaplan, Dhariwal, Neelakantan, Shyam, Sastry, Askell et~al.}]{incontext}
Tom Brown, Benjamin Mann, Nick Ryder, Melanie Subbiah, Jared~D Kaplan, Prafulla Dhariwal, Arvind Neelakantan, Pranav Shyam, Girish Sastry, Amanda Askell, and 1 others. 2020.
\newblock Language models are few-shot learners.
\newblock \emph{Advances in neural information processing systems}, 33:1877--1901.

\bibitem[{Eric et~al.(2019)Eric, Goel, Paul, Kumar, Sethi, Ku, Goyal, Agarwal, Gao, and Hakkani-Tur}]{woz2.1}
Mihail Eric, Rahul Goel, Shachi Paul, Adarsh Kumar, Abhishek Sethi, Peter Ku, Anuj~Kumar Goyal, Sanchit Agarwal, Shuyang Gao, and Dilek Hakkani-Tur. 2019.
\newblock Multiwoz 2.1: A consolidated multi-domain dialogue dataset with state corrections and state tracking baselines.
\newblock \emph{arXiv preprint arXiv:1907.01669}.

\bibitem[{Gwet(2008)}]{gwet}
Kilem~Li Gwet. 2008.
\newblock Computing inter-rater reliability and its variance in the presence of high agreement.
\newblock \emph{British Journal of Mathematical and Statistical Psychology}, 61(1):29--48.

\bibitem[{Heck et~al.(2020)Heck, van Niekerk, Lubis, Geishauser, Lin, Moresi, and Ga{\v{s}}i{\'c}}]{trippy}
Michael Heck, Carel van Niekerk, Nurul Lubis, Christian Geishauser, Hsien-Chin Lin, Marco Moresi, and Milica Ga{\v{s}}i{\'c}. 2020.
\newblock Trippy: A triple copy strategy for value independent neural dialog state tracking.
\newblock \emph{arXiv preprint arXiv:2005.02877}.

\bibitem[{Hrinchuk et~al.(2020)Hrinchuk, Popova, and Ginsburg}]{sc-BERT}
Oleksii Hrinchuk, Mariya Popova, and Boris Ginsburg. 2020.
\newblock \href {https://doi.org/10.1109/ICASSP40776.2020.9053051} {Correction of automatic speech recognition with transformer sequence-to-sequence model}.
\newblock In \emph{ICASSP 2020 - 2020 IEEE International Conference on Acoustics, Speech and Signal Processing (ICASSP)}, pages 7074--7078.

\bibitem[{Huang and Chen(2020)}]{huang2020learning}
Chao-Wei Huang and Yun-Nung Chen. 2020.
\newblock Learning asr-robust contextualized embeddings for spoken language understanding.
\newblock In \emph{ICASSP 2020-2020 IEEE International Conference on Acoustics, Speech and Signal Processing (ICASSP)}, pages 8009--8013. IEEE.

\bibitem[{Jacqmin et~al.(2023)Jacqmin, Druart, Vielzeuf, Rojas-Barahona, Est{\`e}ve, and Favre}]{jacqmin2023olisia}
L{\'e}o Jacqmin, Lucas Druart, Valentin Vielzeuf, Lina~Maria Rojas-Barahona, Yannick Est{\`e}ve, and Beno{\^\i}t Favre. 2023.
\newblock Olisia: a cascade system for spoken dialogue state tracking.
\newblock \emph{arXiv preprint arXiv:2304.11073}.

\bibitem[{Karimi et~al.(2021)Karimi, Rossi, and Prati}]{aeda}
Akbar Karimi, Leonardo Rossi, and Andrea Prati. 2021.
\newblock Aeda: an easier data augmentation technique for text classification.
\newblock \emph{arXiv preprint arXiv:2108.13230}.

\bibitem[{Karpukhin et~al.(2020)Karpukhin, Oguz, Min, Lewis, Wu, Edunov, Chen, and Yih}]{DPR}
Vladimir Karpukhin, Barlas Oguz, Sewon Min, Patrick~SH Lewis, Ledell Wu, Sergey Edunov, Danqi Chen, and Wen-tau Yih. 2020.
\newblock Dense passage retrieval for open-domain question answering.
\newblock In \emph{EMNLP (1)}, pages 6769--6781.

\bibitem[{Kim et~al.(2021)Kim, Liu, Jin, Papangelis, Gopalakrishnan, Hedayatnia, and Hakkani-T{\"u}r}]{kim2021robust}
Seokhwan Kim, Yang Liu, Di~Jin, Alexandros Papangelis, Karthik Gopalakrishnan, Behnam Hedayatnia, and Dilek Hakkani-T{\"u}r. 2021.
\newblock “how robust ru?”: Evaluating task-oriented dialogue systems on spoken conversations.
\newblock In \emph{2021 IEEE Automatic Speech Recognition and Understanding Workshop (ASRU)}, pages 1147--1154. IEEE.

\bibitem[{Kim et~al.(2019)Kim, Yang, Kim, and Lee}]{kim2019efficient}
Sungdong Kim, Sohee Yang, Gyuwan Kim, and Sang-Woo Lee. 2019.
\newblock Efficient dialogue state tracking by selectively overwriting memory.
\newblock \emph{arXiv preprint arXiv:1911.03906}.

\bibitem[{Liang et~al.(2024)Liang, Wang, Song, Hu, Wang, Li, Xiong, and Tang}]{control2}
Xun Liang, Hanyu Wang, Shichao Song, Mengting Hu, Xunzhi Wang, Zhiyu Li, Feiyu Xiong, and Bo~Tang. 2024.
\newblock Controlled text generation for large language model with dynamic attribute graphs.
\newblock \emph{arXiv preprint arXiv:2402.11218}.

\bibitem[{Loshchilov and Hutter(2017)}]{loshchilov2017decoupled}
Ilya Loshchilov and Frank Hutter. 2017.
\newblock Decoupled weight decay regularization.
\newblock \emph{arXiv preprint arXiv:1711.05101}.

\bibitem[{Nechaev et~al.(2021)Nechaev, Ruan, and Kiss}]{Nechaev2021}
Yaroslav Nechaev, Weitong Ruan, and Imre Kiss. 2021.
\newblock \href {https://www.amazon.science/publications/towards-nlu-model-robustness-to-asr-errors-at-scale} {Towards nlu model robustness to asr errors at scale}.
\newblock In \emph{KDD 2021 Workshop on Data-Efficient Machine Learning}.

\bibitem[{Ouyang et~al.(2022)Ouyang, Wu, Jiang, Almeida, Wainwright, Mishkin, Zhang, Agarwal, Slama, Ray, Schulman, Hilton, Kelton, Miller, Simens, Askell, Welinder, Christiano, Leike, and Lowe}]{openai}
Long Ouyang, Jeff Wu, Xu~Jiang, Diogo Almeida, Carroll~L. Wainwright, Pamela Mishkin, Chong Zhang, Sandhini Agarwal, Katarina Slama, Alex Ray, John Schulman, Jacob Hilton, Fraser Kelton, Luke Miller, Maddie Simens, Amanda Askell, Peter Welinder, Paul Christiano, Jan Leike, and Ryan Lowe. 2022.
\newblock \href {https://arxiv.org/abs/2203.02155} {Training language models to follow instructions with human feedback}.
\newblock \emph{Preprint}, arXiv:2203.02155.

\bibitem[{Pal et~al.(2020)Pal, Guillot, Shrivastava, Renders, and Besacier}]{pal2020modeling}
Vaishali Pal, Fabien Guillot, Manish Shrivastava, Jean-Michel Renders, and Laurent Besacier. 2020.
\newblock Modeling asr ambiguity for dialogue state tracking using word confusion networks.
\newblock \emph{arXiv preprint arXiv:2002.00768}.

\bibitem[{Radford et~al.(2022)Radford, Kim, Xu, Brockman, McLeavey, and Sutskever}]{whisper}
Alec Radford, Jong~Wook Kim, Tao Xu, Greg Brockman, Christine McLeavey, and Ilya Sutskever. 2022.
\newblock \href {https://doi.org/10.48550/ARXIV.2212.04356} {Robust speech recognition via large-scale weak supervision}.
\newblock \emph{arXiv preprint}.

\bibitem[{Radford et~al.(2019)Radford, Wu, Child, Luan, Amodei, Sutskever et~al.}]{gpt2}
Alec Radford, Jeffrey Wu, Rewon Child, David Luan, Dario Amodei, Ilya Sutskever, and 1 others. 2019.
\newblock Language models are unsupervised multitask learners.
\newblock \emph{OpenAI blog}, 1(8):9.

\bibitem[{Roberts et~al.(2019)Roberts, Raffel, Lee, Matena, Shazeer, Liu, Narang, Li, and Zhou}]{t5}
Adam Roberts, Colin Raffel, Katherine Lee, Michael Matena, Noam Shazeer, Peter~J Liu, Sharan Narang, Wei Li, and Yanqi Zhou. 2019.
\newblock Exploring the limits of transfer learning with a unified text-to-text transformer.

\bibitem[{Robertson et~al.(2009)Robertson, Zaragoza et~al.}]{bm25}
Stephen Robertson, Hugo Zaragoza, and 1 others. 2009.
\newblock The probabilistic relevance framework: Bm25 and beyond.
\newblock \emph{Foundations and Trends{\textregistered} in Information Retrieval}, 3(4):333--389.

\bibitem[{Sahu et~al.(2023)Sahu, Vechtomova, Bahdanau, and Laradji}]{semaug2}
Gaurav Sahu, Olga Vechtomova, Dzmitry Bahdanau, and Issam~H Laradji. 2023.
\newblock Promptmix: A class boundary augmentation method for large language model distillation.
\newblock \emph{arXiv preprint arXiv:2310.14192}.

\bibitem[{Sennrich et~al.(2015)Sennrich, Haddow, and Birch}]{BT}
Rico Sennrich, Barry Haddow, and Alexandra Birch. 2015.
\newblock Improving neural machine translation models with monolingual data.
\newblock \emph{arXiv preprint arXiv:1511.06709}.

\bibitem[{Sharma et~al.(2020)Sharma, Abraham, Taneja, and Jyothi}]{sharma2020improving}
Yash Sharma, Basil Abraham, Karan Taneja, and Preethi Jyothi. 2020.
\newblock Improving low resource code-switched asr using augmented code-switched tts.
\newblock \emph{arXiv preprint arXiv:2010.05549}.

\bibitem[{Shen et~al.(2018)Shen, Pang, Weiss, Schuster, Jaitly, Yang, Chen, Zhang, Wang, Skerrv-Ryan et~al.}]{taco}
Jonathan Shen, Ruoming Pang, Ron~J Weiss, Mike Schuster, Navdeep Jaitly, Zongheng Yang, Zhifeng Chen, Yu~Zhang, Yuxuan Wang, Rj~Skerrv-Ryan, and 1 others. 2018.
\newblock Natural tts synthesis by conditioning wavenet on mel spectrogram predictions.
\newblock In \emph{2018 IEEE international conference on acoustics, speech and signal processing (ICASSP)}, pages 4779--4783. IEEE.

\bibitem[{Shon et~al.(2022)Shon, Pasad, Wu, Brusco, Artzi, Livescu, and Han}]{asapp}
Suwon Shon, Ankita Pasad, Felix Wu, Pablo Brusco, Yoav Artzi, Karen Livescu, and Kyu~J Han. 2022.
\newblock Slue: New benchmark tasks for spoken language understanding evaluation on natural speech.
\newblock In \emph{ICASSP 2022-2022 IEEE International Conference on Acoustics, Speech and Signal Processing (ICASSP)}, pages 7927--7931. IEEE.

\bibitem[{Soltau et~al.(2022)Soltau, Shafran, Wang, Rastogi, Zhao, Jia, Han, Cao, and Miranda}]{dstc11}
Hagen Soltau, Izhak Shafran, Mingqiu Wang, Abhinav Rastogi, Jeffrey Zhao, Ye~Jia, Wei Han, Yuan Cao, and Aramys Miranda. 2022.
\newblock Speech aware dialog system technology challenge (dstc11).
\newblock \emph{arXiv preprint arXiv:2212.08704}.

\bibitem[{Su et~al.(2021)Su, Shu, Mansimov, Gupta, Cai, Lai, and Zhang}]{pptod}
Yixuan Su, Lei Shu, Elman Mansimov, Arshit Gupta, Deng Cai, Yi-An Lai, and Yi~Zhang. 2021.
\newblock Multi-task pre-training for plug-and-play task-oriented dialogue system.
\newblock \emph{arXiv preprint arXiv:2109.14739}.

\bibitem[{Sun et~al.(2023)Sun, Tian, Zhou, Xu, Hu, Gupta, Wieting, Peng, and Ma}]{control1}
Jiao Sun, Yufei Tian, Wangchunshu Zhou, Nan Xu, Qian Hu, Rahul Gupta, John~Frederick Wieting, Nanyun Peng, and Xuezhe Ma. 2023.
\newblock Evaluating large language models on controlled generation tasks.
\newblock \emph{arXiv preprint arXiv:2310.14542}.

\bibitem[{Touvron et~al.(2023)Touvron, Martin, Stone, Albert, Almahairi, Babaei, Bashlykov, Batra, Bhargava, Bhosale, Bikel, Blecher, Ferrer, Chen, Cucurull, Esiobu, Fernandes, Fu, Fu, Fuller, Gao, Goswami, Goyal, Hartshorn, Hosseini, Hou, Inan, Kardas, Kerkez, Khabsa, Kloumann, Korenev, Koura, Lachaux, Lavril, Lee, Liskovich, Lu, Mao, Martinet, Mihaylov, Mishra, Molybog, Nie, Poulton, Reizenstein, Rungta, Saladi, Schelten, Silva, Smith, Subramanian, Tan, Tang, Taylor, Williams, Kuan, Xu, Yan, Zarov, Zhang, Fan, Kambadur, Narang, Rodriguez, Stojnic, Edunov, and Scialom}]{touvron2023llama}
Hugo Touvron, Louis Martin, Kevin Stone, Peter Albert, Amjad Almahairi, Yasmine Babaei, Nikolay Bashlykov, Soumya Batra, Prajjwal Bhargava, Shruti Bhosale, Dan Bikel, Lukas Blecher, Cristian~Canton Ferrer, Moya Chen, Guillem Cucurull, David Esiobu, Jude Fernandes, Jeremy Fu, Wenyin Fu, and 49 others. 2023.
\newblock \href {https://arxiv.org/abs/2307.09288} {Llama 2: Open foundation and fine-tuned chat models}.
\newblock \emph{Preprint}, arXiv:2307.09288.

\bibitem[{Wei and Zou(2019)}]{eda}
Jason Wei and Kai Zou. 2019.
\newblock Eda: Easy data augmentation techniques for boosting performance on text classification tasks.
\newblock \emph{arXiv preprint arXiv:1901.11196}.

\bibitem[{Whitehouse et~al.(2023)Whitehouse, Choudhury, and Aji}]{semaug1}
Chenxi Whitehouse, Monojit Choudhury, and Alham~Fikri Aji. 2023.
\newblock Llm-powered data augmentation for enhanced crosslingual performance.
\newblock \emph{arXiv preprint arXiv:2305.14288}.

\bibitem[{Yoon et~al.(2023)Yoon, Hwang, Han, Bang, and Kim}]{yoon2023adapting}
Jaeseok Yoon, Seunghyun Hwang, Ran Han, Jeonguk Bang, and Kee-Eung Kim. 2023.
\newblock Adapting text-based dialogue state tracker for spoken dialogues.
\newblock \emph{arXiv preprint arXiv:2308.15053}.

\bibitem[{Young et~al.(2013)Young, Ga{\v{s}}i{\'c}, Thomson, and Williams}]{young2013pomdp}
Steve Young, Milica Ga{\v{s}}i{\'c}, Blaise Thomson, and Jason~D Williams. 2013.
\newblock Pomdp-based statistical spoken dialog systems: A review.
\newblock \emph{Proceedings of the IEEE}, 101(5):1160--1179.

\bibitem[{Zhang et~al.(2021)Zhang, Yi, Tian, Bai, Tao, Liu, and Wen}]{sc-kfold}
Shuai Zhang, Jiangyan Yi, Zhengkun Tian, Ye~Bai, Jianhua Tao, Xuefei Liu, and Zhengqi Wen. 2021.
\newblock \href {https://doi.org/10.21437/Interspeech.2021-1242} {{End-to-End Spelling Correction Conditioned on Acoustic Feature for Code-Switching Speech Recognition}}.
\newblock In \emph{Proc. Interspeech 2021}, pages 266--270.

\bibitem[{Zhang et~al.(2022)Zhang, Roller, Goyal, Artetxe, Chen, Chen, Dewan, Diab, Li, Lin, Mihaylov, Ott, Shleifer, Shuster, Simig, Koura, Sridhar, Wang, and Zettlemoyer}]{zhang2022opt}
Susan Zhang, Stephen Roller, Naman Goyal, Mikel Artetxe, Moya Chen, Shuohui Chen, Christopher Dewan, Mona Diab, Xian Li, Xi~Victoria Lin, Todor Mihaylov, Myle Ott, Sam Shleifer, Kurt Shuster, Daniel Simig, Punit~Singh Koura, Anjali Sridhar, Tianlu Wang, and Luke Zettlemoyer. 2022.
\newblock \href {https://arxiv.org/abs/2205.01068} {Opt: Open pre-trained transformer language models}.
\newblock \emph{Preprint}, arXiv:2205.01068.

\end{thebibliography}
